\documentclass{edm_article}
\usepackage{multirow}
\usepackage{makecell} 
\usepackage[linesnumbered,ruled,vlined]{algorithm2e}
\SetKwBlock{DoParallel}{do $l \gets 1$ \KwTo $L$ in parallel}{end}
\usepackage{hyperref}
\usepackage{soul}
\usepackage{tabularx}
\usepackage{longtable}
\usepackage{tikz}
\usepackage{xcolor}
\definecolor{keywords}{RGB}{0,0,255}
\definecolor{comments}{RGB}{0,128,0}
\definecolor{strings}{RGB}{255,0,0}
\usepackage{listings}
\usepackage{bbm}
\usepackage{ulem, soul}
\usepackage{colortbl} 
\usepackage{natbib}
\usepackage{booktabs}

\usepackage{ulem}
\usepackage[most]{tcolorbox}
\tcbset{
    mybox/.style={
        colback=gray!10,    
        colframe=gray!50,   
        fonttitle=\bfseries, 
        boxrule=0.8mm,      
        title=#1            
    }
}

\begin{document}

\title{Enhancing LLM-Based Short Answer Grading with Retrieval-Augmented Generation}


%
%
%
%

\numberofauthors{9} 
\author{
\alignauthor
Yucheng Chu\\
       \affaddr{Michigan State University}\\
       \email{chuyuch2@msu.edu}
\alignauthor
Peng He\\
       \affaddr{Washington State University}\\
       \email{peng.he@wsu.edu}
\alignauthor 
Hang Li\\
        \affaddr{Michigan State University}\\
       \email{lihang4@msu.edu}
\and  
\alignauthor 
Haoyu Han\\
       \affaddr{Michigan State University}\\
       \email{lhanhaoy1@msu.edu}
\alignauthor Kaiqi Yang\\
       \affaddr{Michigan State University}\\
       \email{kqyang@msu.edu}
\alignauthor Yu Xue\\
       \affaddr{Washington State University}\\
       \email{yu.xue@wsu.edu}
\and
\alignauthor Tingting Li\\
       \affaddr{Washington State University}\\
       \email{tingting.li1@wsu.edu}
\alignauthor Joseph Krajcik\\
       \affaddr{Michigan State University}\\
       \email{krajcik@msu.edu}
\alignauthor Jiliang Tang\\
       \affaddr{Michigan State University}\\
       \email{tangjili@msu.edu}
}


\maketitle

\begin{abstract}
Short answer assessment is a vital component of science education, allowing evaluation of students' complex three-dimensional understanding. Large language models (LLMs) that possess human-like ability in linguistic tasks are increasingly popular in assisting human graders to reduce their workload. However, LLMs' limitations in domain knowledge restrict their understanding in task-specific requirements and hinder their ability to achieve satisfactory performance. Retrieval-augmented generation (RAG) emerges as a promising solution by enabling LLMs to access relevant domain-specific knowledge during assessment. In this work, we propose an adaptive RAG framework for automated grading that dynamically retrieves and incorporates domain-specific knowledge based on the question and student answer context. Our approach combines semantic search and curated educational sources to retrieve valuable reference materials. Experimental results in a science education dataset demonstrate that our system achieves an improvement in grading accuracy compared to baseline LLM approaches. The findings suggest that RAG-enhanced grading systems can serve as reliable support with efficient performance gains.

\end{abstract}


\section{Introduction}

Assessment and analysis of student understanding in science education extend far beyond simple grading. With the emergence of new frameworks for K-12 science education, there is an increasing emphasis on analyzing students' multidimensional comprehension of scientific understanding. The National Research Council's Framework for K-12 Science Education~\cite{national2012framework} established three critical dimensions of science learning: disciplinary core ideas (DCIs), science and engineering practices (SEPs), and crosscutting concepts (CCCs). To demonstrate their achievement, students should apply their 3D understanding to interpret compelling phenomena or solve real-world problems. This framework has fundamentally transformed how we assess student learning, shifting away from traditional multiple-choice questions toward short-answer assessments that better capture students' authentic understanding~\cite{he2023predicting}.
Short-answer questions have become particularly valuable in this context, as they require students to demonstrate their three-dimensional knowledge by explaining real-world phenomena and solving complex problems~\cite{he2023applying}. These responses provide rich insights into students' conceptual understanding, revealing both their mastery and misunderstanding, which is essential for teachers to adapt their instructional strategies and support students' self-regulated learning~\cite{he2024design}. 

However, the analysis of these short answers presents significant challenges as it is time-intensive, requires deep expertise across all three dimensions, and makes it difficult for teachers to provide timely, actionable feedback to individual students. Fortunately, artificial intelligence (AI), particularly automatic short answer grading (ASAG) systems, has emerged as a promising solution to these challenges. Traditional ASAG approaches using machine learning (ML) techniques have shown success in providing consistent and objective scoring but are limited by their reliance on large training datasets~\cite{burrows2015eras, sultan2016fast}. The recent advent of large language models (LLMs) has opened new possibilities for training ASAG systems on smaller datasets, due to its possession of broad general knowledge from pre-training. However, current LLM-based approaches face two critical limitations in science education assessment. First, they exhibit deceptive performance in technical domains, often generating plausible-sounding but scientifically false evaluations when in-depth domain knowledge is required. Their output can be unreliable when evaluating answers that contain ambiguous terminology or complex scientific concepts. Second, they lack sufficient task-specific knowledge about the grading criteria~\cite{lewis2020retrieval}. They often fail to understand the nuanced requirements of specific educational frameworks such as the three-dimensional learning approach, which can lead to misalignment between automated scores and expert evaluations. 

To address these limitations, we introduce GradeRAG, a novel framework that enhances LLM-based grading through retrieval-augmented generation (RAG). 
Our approach implements a specialized RAG pipeline that provides the LLM with access to curated domain-specific knowledge bases, enabling a more accurate assessment of scientific concepts across the three dimensions (DCIs, SEPs, and CCCs). This knowledge-focused approach is particularly crucial in science education, where effective assessment requires domain-specific expertise of specialized concepts and their interconnections~\cite{he2023applying, beatty2014developing}.
As part of our knowledge retrieval strategy, we incorporate expert-annotated scoring rationales as a specialized form of knowledge source. These examples contain detailed scoring rationales that explicitly identify critical scientific terminology and reasoning patterns in student responses. By treating these examples as retrievable knowledge, LLMs are guided to emulate expert analysis processes, further improving scoring performance.
We evaluate GradeRAG on a dataset of student short answers on science assessments. Results demonstrate improvements in both grading accuracy and consistency compared to the baseline approaches. Our results suggest that integrating specialized knowledge retrieval systems can bridge the gap between automated efficiency and expert-level assessment in science education, leading to more reliable evaluation of complex scientific understanding.

\section{Related Work}
\vspace{-0.1in}
\paragraph{Automatic short answer grading}
Automatic short answer grading (ASAG) has evolved significantly over the past decades. Early approaches mainly focus on text matching and statistics-based methods~\cite{mohler2011learning}. With the advance in ML, later systems employ feature engineering techniques with supervised training methods~\cite{leacock2003c-rater}. These traditional approaches typically require large training datasets and often struggle with capturing the semantic differences in student answers. Recent works utilizing neural networks demonstrate improvement in understanding semantic relationships in short answers~\cite{hassan2018automatic}. Large language models (LLMs) further enable enhanced capabilities in ASAG through zero-shot and few-shot~\cite{lee2024applyingllm} techniques. However, as noted by~\cite{li2023adapting}, LLMs still face challenges in specialized domains like science education, where domain-specific knowledge is crucial for accurate assessment.

\vspace{-0.2in}
\paragraph{Retrieval-augmented generation} Retrieval-augmented generation (RAG) has emerged as an effective approach to enhance LLMs' reasoning ability by providing access to external knowledge sources. RAG has demonstrated effectiveness in knowledge-intensive tasks requiring factual accuracy~\cite{lewis2020retrieval} and domain expertise~\cite{guu2020retrieval}. In the educational context, the application of RAG remains relatively unexplored, with most works focusing on content generation~\cite{miladi2024comparative} rather than assessment. Recent studies have begun investigating RAG for automated assessment. For example, ~\cite{harshavardhan2024rubric} proposes a rubric-centric approach for automated test correction using RAG, while ~\cite{sundar2024revolutionizing} evaluated RAG frameworks for grading open-ended written responses. However, the application of RAG, specifically for short answer grading in science education, where domain-specific knowledge across multiple dimensions (DCIs, SEPs, CCCs) is crucial, represents a novel contribution to our work.

\section{Method}
\subsection{Problem Statement}

The task of automatic short answer grading (ASAG) involves assigning appropriate scores to open-ended student responses. Given a student answer $x$, an ASAG system maps it to one of $C$ predefined score levels $\{s_1, ..., s_C\}$. Formally, an ASAG system $\mathcal{F}$ generates score predictions $\hat{y}_i = \mathcal{F}(x_i)$ for each response $x_i$ across each dimension.
This work extends traditional ASAG through retrieval-augmented generation. We leverage two types of external knowledge resources to enhance grading performance. First, we utilize domain-specific knowledge from an external database $K_D$, retrieving relevant background information $I$ that contains educational materials and scoring guidelines. Second, we incorporate a specialized knowledge collection $K_E$ of $N$ expert-annotated graded examples $\mathcal{E}=\{(x_i, r_i, y_i)|i=1,...,N\}$, where each entry consists of a student response $x_i$, a scoring rationale $r_i$, and the corresponding ground truth score $y_i$. By combining these knowledge sources, our enhanced ASAG system $\mathcal{F}(x_i, I, \mathcal{E})$ creates a more comprehensive context for LLM-as-a-Grader's accurate assessment.

\subsection{RAG-Based Grading System\label{sec:rag_framework}}

In this section, we present GradeRAG, a framework that combines in-context learning with RAG for automatic short answer grading in science education. 

\begin{figure}[htbp]
\Description{Illustration of the proposed framework}  
\centering

\includegraphics[width=0.475\textwidth]{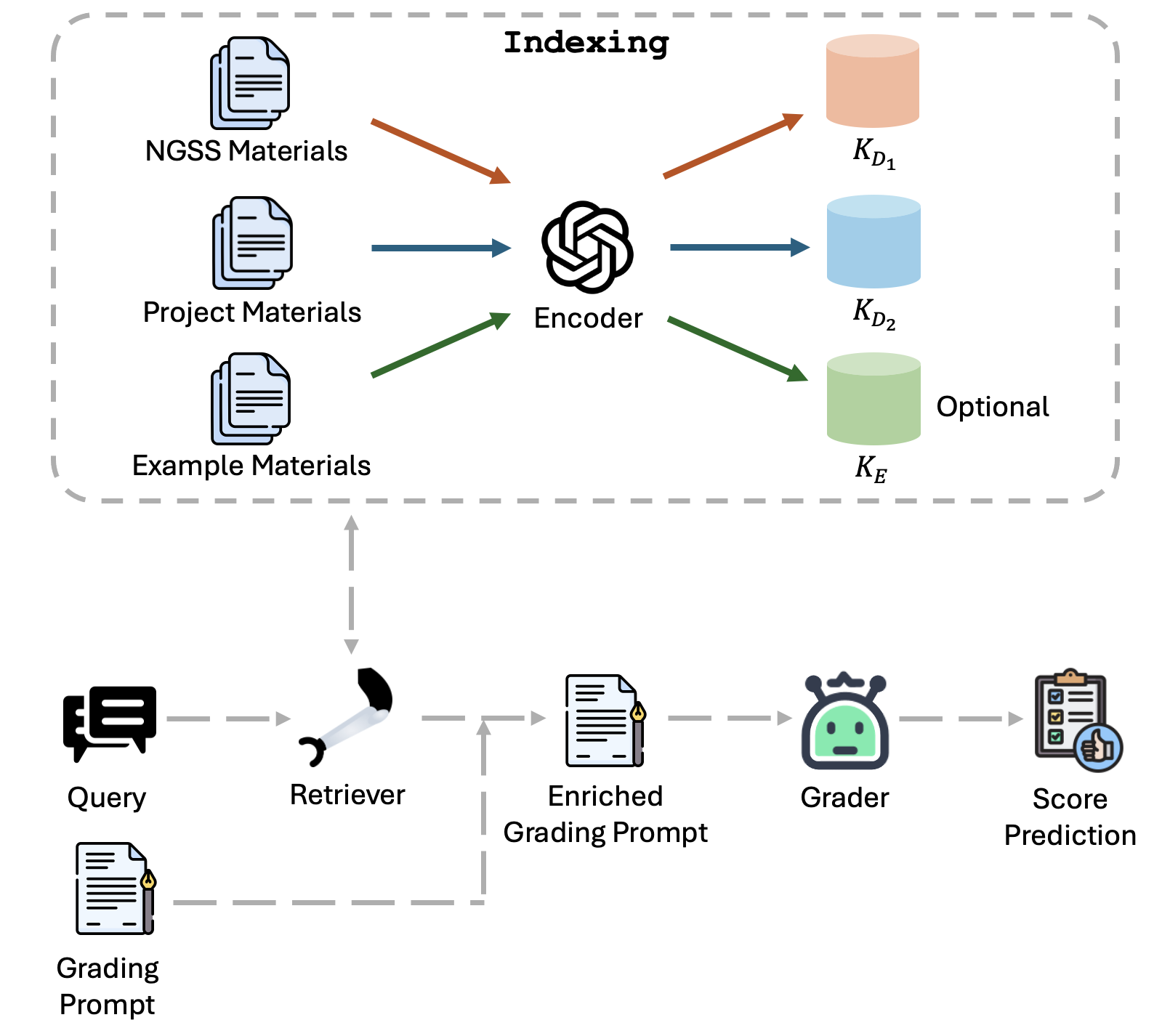} 
\caption{An illustration of the proposed GradeRAG framework.}
\label{fig:framework_illustration}
\end{figure}

\subsubsection{Knowledge Base Construction}

As shown in Figure ~\ref{fig:framework_illustration}, our system constructs a triple-index knowledge base optimized for scientific assessment. $K_{D_1}$ and $K_{D_2}$ consists of project-related information and grading guidelines, while $K_{E}$ (which is optinal) contains expert-annotated scoring examples. The first index $K_{D_1}$ contains general assessment documents such as the Framework of K-12 Science Education~\cite{national2012framework} and the NGSS standard document~\cite{ngss2013next}. These materials are systematically chunked into coherent segments of approximately 512 tokens, which balances content completeness with the retrieval granularity. This chunking size preserves sufficient context while enabling precise retrieval of relevant concepts, as smaller chunks would fragment scientific explanations while larger chunks can introduce excessive irrelevant information. Each chunk is tagged with metadata including the document source and dimension type to facilitate efficient filtering during retrieval. 

The second index $K_{D_2}$ consists of task-specific materials, such as the three-dimensional learning progression materials~\cite{he2023applying}, which includes unpacking DCIs, SEPs, and CCCs materials and the holistic levels in the three assessment tasks. For these more specialized resources, we employ a finer-grained, manual chunking strategy. Specifically, we divide the content into single sentences or meaningful segments containing one or two bullet points. This approach is adopted as task-specific materials contain dense, concentrated information where even small segments can provide critical assessment criteria. With this fine-grained chunking strategy, our system can retrieve precisely the relevant assessment standards without including extraneous information. 

Each chunk is tagged with metadata indicating the dimension type, level number, and chunk identifier to enable targeted filtering during retrieval.

The third index $K_E$ comprises expert-annotated examples, each encoded with metadata such as score assignments and dimension-specific annotations. These examples are maintained as complete units rather than being chunked, which preserves the complete reasoning logic of expert scoring. This design choice ensures that our framework can access the full scoring demonstration with coherent guidance for emulating human-like reasoning processes, thus improving the alignment between model scores and human evaluation.


\subsubsection{Dual Retrieval Strategy}

For each student response $x$, GradeRAG employs a dual retrieval mechanism to gather complementary information. 


\paragraph{Knowledge Retrieval} The system first identifies relevant dimension-specific content using a task-aware retrieval strategy. Each task is associated with specific dimension levels (e.g., Task 1 corresponds to DCI Level 1 and 2, while Task 2 encompasses Level 4, 5 and 6).
This retrieval mechanism operates in two steps. First, it retrieves dimension-specific materials using a semantic similarity search over the indexed content that corresponds to different levels. The top-$k$ pertinent information is retrieved. Second, to optimize relevance to the student answer while accounting for the context window constraint, we implement a reranking mechanism that combines three weighted components: semantic similarity (40\% weight), text matching based on word overlap (30\% weight), and domain-specific concept coverage (30\% weight). 
We deliberately combine these three scores to address the different aspects of information relevance in science education assessment. Semantic similarity captures contextual relevance and identifies materials that share similar meanings. Text matching based on normalized word overlap between the student response and knowledge chunks can identify the lexical connections. This ensures that specific terminology in student answers is represented in the retrieved content. Lastly, the concept matching score evaluates the coverage of key scientific concepts necessary for assessment, such as properties, substances, chemical reactions, and identifications in the knowledge chunk. The final top-$k$ passages are selected based on this combined score. By weighting and combining these three signals, our system achieves more comprehensive retrieval.


\paragraph{Example Retrieval}
In parallel with knowledge retrieval, our system retrieves similar expert-annotated examples using semantic similarity matching. These examples include expert rationales that explain the scoring criteria and highlight the critical keywords in student responses that signal understanding. 

Figure~\ref{fig:expert_rationale_example} shows a concrete example of the expert-annotated grading sample. We incorporate demonstrations (human-graded examples) into the grading prompt using a scale-based approach. The number of demonstrations scales with the complexity of the scoring scheme. For binary classification tasks with two score classes, we utilize six demonstrations, while for ternary classification tasks with three score classes nine demonstrations are employed. This scaling ensures sufficient examples across all possible score categories while maintaining prompt efficiency.

\begin{figure}[!btph]
\begin{tcolorbox}[mybox={Examplar of Expert-Annotated Sample with Scoring Rationale (SEP)}]

\textbf{Student's Short Answer: }
When coconut oil is mixed with lye, a chemical reaction occurs because soap and glycerol are new substances. From the table, I found that the odor, density, solubility in water, and melting point are different from each other. They are properties that can be used to identify substances and whether a chemical reaction occurs. \newline

\textbf{Score:} 
SEP-1 \newline

\textbf{Scoring Rationale: }
``When coconut oil is mixed with lye, a chemical reaction occurs because soap and glycerol are new substances.'' – this part meets the \textit{partial} SEP criteria that mentioned a descriptive explanation, including a claim of a chemical reaction occurs and observed evidence of the data before and after the process. However, the response \textit{did not use the evidence} to connect to the phenomenon – new substance produced and a chemical reaction occurred.

\end{tcolorbox}
\caption{An example of expert-elaborated rationales for a short-answer sample on the SEP dimension.}
\label{fig:expert_rationale_example}
\end{figure}

\subsubsection{Grading}
After retrieving relevant knowledge and examples, GradeRAG performs the grading process. 
For each student response, GradeRAG assembles a unified grading context by combining the original expert-designed task guidelines, the retrieved top-$k$ knowledge chunks, and similar expert-annotated examples with their rationales. This comprehensive content combination augments the expert-designed base grading guideline, providing additional context and criteria for more accurate assessment. The combined prompt (shown in Figure~\ref{fig:grader_prompt}) is passed to the LLM-powered Grader agent for score generation. Formally, the Grader generates 
$\hat{y}_{1:T} = \mathcal{F}(x, K, E)$
where $\hat{y}_{1:T-1}$ represents the reasoning rationale containing $T-1$ tokens that explain the grading decision, and $\hat{y}_{T}$ is the final token that represents the numerical score. The input consists of the student answer $x$, retrieved knowledge chunks $K$, and expert examples $E$.
The scoring process is performed separately for each dimension to ensure a focused evaluation of the different aspects of scientific understanding. The Grader's output includes both a numerical score $\hat{y}_{T}$ indicating the student's achievement level and a detailed explanation $\hat{y}_{1:T-1}$ that references the specific evidence from the student response and connects it to the retrieved knowledge. This approach simulates the expert annotation process to enhance grading alignment.

\begin{figure}[!btph]
\begin{tcolorbox}[mybox={Prompt for Grader}]

\textbf{Task: }
<Task Description>

\textbf{Question:}
Write a scientific explanation about whether a chemical reaction occurs in the described scenario. Make sure your explanation includes a claim, evidence, and reasoning. \newline

\textbf{Step 1}: Review and apply the following detailed learning goals and scoring criteria for grading student answer:

$\bullet$ \textbf{Criteria:}
<Initial Expert Criteria>

$\bullet$ \textbf{Knowledge Materials:}
<Retrieved Knowledge> \newline

\textbf{Step 2}: Examine following example graded answers. Analyze how each one is assessed as explained in scoring rationales: 
<Retrieved Examples> \newline

\textbf{Step 3}: Now assess the fulfillment of student answer based on the description of task, question, criteria, and graded examples. Give a score of 0, 1, or 2 with your reasoning:
<Student Answer>

\end{tcolorbox}
\caption{An exemplary prompt for grader}
\label{fig:grader_prompt}
\end{figure}

\section{Experiment}

This section presents our experimental evaluation for our GradeRAG system. Specifically, we intend to investigate whether incorporating external knowledge through RAG improves automated scoring performance compared to the baseline approaches. We further investigate whether GradeRAG is compatible with \textit{in-context learning} by conducting an ablation study.


\subsection{Dataset}

Our evaluation uses a testing dataset $\mathcal{D}_{test}$ comprising student responses collected from two middle schools in the Midwestern United States. The dataset includes student responses for each of the three NGSS-aligned science tasks, completed in standard classroom settings. The detailed response count for $\mathcal{D}_{test}$ is reported in Table~\ref{tab:dataset_detail_table}. To establish ground truth scores, two content experts independently graded all responses using standardized rubrics and demonstration examples. Any scoring discrepancies were resolved through expert discussion to ensure high-quality gold standard annotations. Table~\ref{tab:dataset_detail_table} shows the details of level coverage for each question task.



\begin{table}[!btph]
\centering
\caption{Detailed statistics of different questions in dataset $\mathcal{D}_{test}$. The number of samples in each label category is shown as $C_i$.}
\label{tab:dataset_detail_table}
\renewcommand{\arraystretch}{1.05}
\resizebox{0.425\textwidth}{!}{
\begin{tabular}{@{}cll|cc@{}}
\toprule
\multirow{2}{*}{\textbf{Question}} & \multicolumn{2}{c|}{\multirow{2}{*}{\textbf{Assessment Level}}} & \multicolumn{2}{c}{\textbf{Response Count}}  \\ \cmidrule{4-5}
& & & \textbf{Total} & C1 / C2 / C3 \\ \midrule
\multirow{3}{*}{$Q_1$} & \cellcolor[HTML]{EFEFEF}DCI & (0, 1, 2) & \multirow{3}{*}{\cellcolor[HTML]{EFEFEF}} & 28/14/5   \\
& \cellcolor[HTML]{EFEFEF}SEP & (0, 1, 2) & \multirow{1}{*}{\cellcolor[HTML]{EFEFEF}47}  & 44/3/0  \\
& \cellcolor[HTML]{EFEFEF}CCC & (0, 1, 2) & \multirow{1}{*}{\cellcolor[HTML]{EFEFEF}} &  44/3/0 \\ \midrule
\multirow{3}{*}{$Q_2$} & \cellcolor[HTML]{EFEFEF}DCI &  (0, 4, 5, 6) & \cellcolor[HTML]{EFEFEF}\multirow{3}{*}{} & 13/15/1/2 \\
 & \cellcolor[HTML]{EFEFEF}SEP & (0, 1, 2) & \cellcolor[HTML]{EFEFEF}\multirow{1}{*}{31} & 12/3/16 \\
& \cellcolor[HTML]{EFEFEF}CCC & (0, 1, 2) & \cellcolor[HTML]{EFEFEF}\multirow{3}{*}{}  & 10/18/3 \\ \midrule
\multirow{3}{*}{$Q_3$} & \cellcolor[HTML]{EFEFEF}DCI & (0, 6, 7) & \cellcolor[HTML]{EFEFEF}\multirow{1}{*}{} & 40/3/3 \\ 
& \cellcolor[HTML]{EFEFEF}SEP & (0, 1, 2, 3) & \cellcolor[HTML]{EFEFEF}\multirow{1}{*}{46} & 12/26/5/3 \\
 & \cellcolor[HTML]{EFEFEF}CCC & (0, 1, 2, 3) &  \cellcolor[HTML]{EFEFEF}\multirow{1}{*}{} & 14/1/28/3 \\ \bottomrule
\end{tabular}}
\end{table}


\subsection{Experimental Setting}
We implement our grading system using gpt-4o-mini-2024-07-18 as the core Grader agent. To ensure reproducibility and consistent evaluation, we set the temperature parameter to 0, eliminating stochastic variations in the model's output. The embedding model used for indexing documents is text-embedding-ada-002. Since the original student answers were collected in image format, we first process them using gpt-4-vision-preview for transcription. These transcribed responses are then passed to the Retriever and Grader agent for separately assessing each task dimension (e.g., DCI, SEP, CCC). 
We run experiments for testing our system under two experimental conditions: NonRAG and GradeRAG. Under naive prompt, the system processes student responses using only the basic grading rubric without additional context or examples from the RAG component. Under GradeRAG with zero-shot prompting, the system incorporates the retrieved materials as described in Section~\ref{sec:rag_framework}, but without expert-annotated examples. 

We evaluate performance using standard metrics, including accuracy, weighted F1-score, and Cohen's kappa. To be specific, 
Accuracy ($Acc$) measures the proportion of correct predictions. Weighted F1-score ($\text{F1}_{\text{weighted}}$) accounts for class imbalance by computing the F1-score for each class and taking their weighted average. 
Cohen's kappa ($\kappa$) measures the inter-rater reliability by accounting for the possibility of agreement occurring by chance. The formulas for these metrics are as follows:
\begin{align}\nonumber
    &Acc = \frac{1}{N}\sum_{i=1}^N \mathbb{I}_{(y_i=\hat{y}_i)},\ \kappa = \frac{p_o - p_e}{1 - p_e},\\ \nonumber&\text{F1}_{\text{weighted}} = \sum{k=1}^K n_k \cdot \frac{2 \cdot P_k \cdot R_k}{P_k + R_k}
\end{align}

where $p_o$ is the observed agreement between the model and expert scores, $p_e$ is the expected agreement by chance, $n_k$ is the proportion of samples in class $k$, $P_k$ and $R_k$ are the precision and recall for class $k$.


\vspace{-3mm}
\begin{table}[!btph]
\centering
\caption{Performance comparison between NonRAG and GradeRAG on $\mathcal{D}_{test}$ with zero-shot prompting (shot number $C$=0, retrieval size $k$=4).}
\label{tab:result_rag}
\renewcommand{\arraystretch}{1.05}
\resizebox{.455\textwidth}{!}{
\begin{tabular}{@{}ccccccc@{}}
\toprule
\multirow{2}{*}{\textbf{Question}}  & \multicolumn{3}{c|}{\textbf{NonRAG}} & \multicolumn{3}{c}{\textbf{GradeRAG}} \\ \cmidrule{2-7}
& \multicolumn{1}{c}{DCI} & SEP & \multicolumn{1}{c|}{CCC} & DCI & SEP & CCC \\ \midrule
\multicolumn{7}{c}{\textbf{Accuracy} (Acc)} \\ \midrule
\rowcolor[HTML]{EFEFEF}\multicolumn{1}{c|}{$Q_1$}  & 0.217  &  0.717 &   \multicolumn{1}{c|}{0.739 }  & 0.348 & 0.804 & 0.957 \\
\multicolumn{1}{c|}{$Q_2$} & 0.355 &  0.387 &    \multicolumn{1}{c|}{0.581}  & 0.484  & 0.419 &  0.645 \\
\rowcolor[HTML]{EFEFEF}\multicolumn{1}{c|}{$Q_3$}  & 0.717 & 0.478 & \multicolumn{1}{c|}{0.565}  & 0.783  & 0.522 & 0.674  \\ \midrule
\multicolumn{7}{c}{\textbf{Weighted F1 Score} (F1)} \\ \midrule
\rowcolor[HTML]{EFEFEF}\multicolumn{1}{c|}{$Q_1$} & 0.170 &  0.799 &  \multicolumn{1}{c|}{0.816}  & 0.257 & 0.858 &  0.935   \\
\multicolumn{1}{c|}{$Q_2$}  & 0.399  & 0.463 &  \multicolumn{1}{c|}{0.544}  & 0.527 & 0.499 & 0.595 \\
\rowcolor[HTML]{EFEFEF}\multicolumn{1}{c|}{$Q_3$}  & 0.752 &  0.443 &  \multicolumn{1}{c|}{0.543}  & 0.792 &  0.481 &  0.656   \\ \midrule
\multicolumn{7}{c}{\textbf{Cohen's Kappa} ($\kappa$)} \\ \midrule
\rowcolor[HTML]{EFEFEF}\multicolumn{1}{c|}{$Q_1$} & -0.001 & 0.188 &  \multicolumn{1}{c|}{0.074}  & 0.100  & 0.119 &  0.000  \\
\multicolumn{1}{c|}{$Q_2$}  & 0.129 & 0.232  & \multicolumn{1}{c|}{0.188}  &  0.245 & 0.260 &   0.288   \\
\rowcolor[HTML]{EFEFEF}\multicolumn{1}{c|}{$Q_3$}  & 0.124 & 0.083 &    \multicolumn{1}{c|}{0.123}  &  0.196 & 0.174 & 0.331  \\ \bottomrule
\end{tabular}}
\end{table}

\begin{table}[!btph]
\centering
\caption{Performance comparison between NonRAG and GradeRAG under in-context learning (ICL) on $\mathcal{D}_{test}$.}
\label{tab:ablation_ICL}
\renewcommand{\arraystretch}{1.05}
\resizebox{.475\textwidth}{!}{
\begin{tabular}{@{}ccccccc@{}}
\toprule
\multirow{2}{*}{\textbf{Question}}  & \multicolumn{3}{c|}{\textbf{NonRAG-ICL}} &  \multicolumn{3}{c}{\textbf{GradeRAG-ICL}}  \\ \cmidrule{2-7}
& \multicolumn{1}{c}{DCI} & SEP & \multicolumn{1}{c|}{CCC} & DCI & SEP & CCC \\ \midrule
\multicolumn{7}{c}{\textbf{shot number $C$ = 3, retrieval size $k$ = 1}} \\ \hline
\multicolumn{7}{c}{\textbf{Accuracy} (Acc)} \\ \midrule
\rowcolor[HTML]{EFEFEF}\multicolumn{1}{c|}{$Q_1$}  & 0.609 & 0.543 &  \multicolumn{1}{c|}{0.826}  &  0.609  & 0.565 &  0.913 \\
\multicolumn{1}{c|}{$Q_2$} &  0.742  & 0.226  & \multicolumn{1}{c|}{0.387} &  0.742  & 0.323 & 0.548\\
\rowcolor[HTML]{EFEFEF}\multicolumn{1}{c|}{$Q_3$}  &  0.674 & 0.500 & \multicolumn{1}{c|}{0.304}  &  0.696  & 0.587 &  0.348 \\ \midrule
\multicolumn{7}{c}{\textbf{Weighted F1 Score} (F1)} \\ \midrule
\rowcolor[HTML]{EFEFEF}\multicolumn{1}{c|}{$Q_1$} & 0.591 & 0.664  &  \multicolumn{1}{c|}{0.875}   & 0.599  & 0.682 & 0.927   \\
\multicolumn{1}{c|}{$Q_2$}  & 0.717 & 0.229 &  \multicolumn{1}{c|}{0.315}  &  0.742  & 0.340  &  0.537 \\
\rowcolor[HTML]{EFEFEF}\multicolumn{1}{c|}{$Q_3$}  & 0.726 & 0.473 &  \multicolumn{1}{c|}{0.300}  & 0.741 & 0.555  &  0.327   \\ \midrule
\multicolumn{7}{c}{\textbf{Cohen's Kappa} ($\kappa$)} \\ \midrule
\rowcolor[HTML]{EFEFEF}\multicolumn{1}{c|}{$Q_1$} & 0.230  & 0.087 &  \multicolumn{1}{c|}{0.281}    & 0.303 &  0.094  &  0.292  \\
\multicolumn{1}{c|}{$Q_2$} & 0.536 & 0.055 & \multicolumn{1}{c|}{0.394}  &  0.554  & 0.187  & 0.259     \\
\rowcolor[HTML]{EFEFEF}\multicolumn{1}{c|}{$Q_3$}  & 0.170 & 0.108  &    \multicolumn{1}{c|}{0.142}  &  0.189 & 0.231 & 0.142  \\ \bottomrule

\multicolumn{7}{c}{\textbf{shot number $C$ = 6, retrieval size $k$ = 2}} \\ \hline
\multicolumn{7}{c}{\textbf{Accuracy} (Acc)} \\ \midrule
\rowcolor[HTML]{EFEFEF}\multicolumn{1}{c|}{$Q_1$}  &  0.609 & 0.696 & \multicolumn{1}{c|}{0.891} & 0.630 & 0.717 & 0.913 \\
\multicolumn{1}{c|}{$Q_2$} &  0.677 & 0.290 &  \multicolumn{1}{c|}{0.419}  & 0.774 & 0.355 & 0.419  \\
\rowcolor[HTML]{EFEFEF}\multicolumn{1}{c|}{$Q_3$}  & 0.761 & 0.413  & \multicolumn{1}{c|}{0.391}  & 0.761  &  0.413 & 0.348 \\ \midrule
\multicolumn{7}{c}{\textbf{Weighted F1 Score} (F1)} \\ \midrule
\rowcolor[HTML]{EFEFEF}\multicolumn{1}{c|}{$Q_1$} & 0.584 & 0.786 &  \multicolumn{1}{c|}{0.913}  &  0.609 & 0.798 & 0.927  \\
\multicolumn{1}{c|}{$Q_2$}  &  0.665  & 0.320 &  \multicolumn{1}{c|}{0.333}  & 0.758 & 0.395  & 0.333 \\
\rowcolor[HTML]{EFEFEF}\multicolumn{1}{c|}{$Q_3$}  & 0.785 & 0.400 &  \multicolumn{1}{c|}{0.317}  & 0.784 & 0.392  &   0.272  \\ \midrule
\multicolumn{7}{c}{\textbf{Cohen's Kappa} ($\kappa$)} \\ \midrule
\rowcolor[HTML]{EFEFEF}\multicolumn{1}{c|}{$Q_1$} & 0.210 & 0.053 &  \multicolumn{1}{c|}{0.238} & 0.276 & 0.178 &  0.292  \\
\multicolumn{1}{c|}{$Q_2$}  &  0.434 & 0.139  & \multicolumn{1}{c|}{0.120}  &  0.600 &  0.213 &   0.120   \\
\rowcolor[HTML]{EFEFEF}\multicolumn{1}{c|}{$Q_3$}  & 0.259 & -0.011 &    \multicolumn{1}{c|}{0.129}  & 0.259  & -0.006 &  0.068 \\ \bottomrule

\end{tabular}}
\end{table}

\subsection{Main Results}

Table~\ref{tab:result_rag} presents the comparative performance of our proposed GradeRAG framework against the non-retrieval baseline (NonRAG) across three questions ($Q_1$, $Q_2$, and $Q_3$) and three assessment dimensions (DCI, SEP, and CCC). The results demonstrate general performance improvements with GradeRAG across metrics, questions, and dimensions. 
For all three metrics (accuracy, weighted F1 score, and Cohen's Kappa), GradeRAG outperforms the non-retrieval baseline in most cases. The most substantial gains are observed in $Q_1$ for the DCI dimension (13.1\% accuracy increase) and CCC dimension (21.8\% accuracy increase), with $Q_3$'s CCC dimension showing the largest Kappa improvement (20.8\% increase). While the Kappa values remain relatively low overall (ranging from 0.000 to 0.331), GradeRAG generally achieves better alignment with expert grading patterns compared to NonRAG.

Analyzing performance across questions reveals patterns related to question complexity. $Q_1$, which has a relatively simple assessment structure with three levels (0, 1, 2) across all dimensions and highly imbalanced class distributions, demonstrates the largest average performance gains (14.5\% in accuracy). This suggests that for questions with clear-cut scoring criteria and imbalanced response distributions, retrieved knowledge can provide particularly effective guidance. In contrast, $Q_2$ and $Q_3$ involve more complex assessment criteria with up to four score levels and more balanced class distributions, resulting in more moderate performance improvements. In general, this variety in complexity levels across questions suggests how retrieval effectiveness may depend on both the nature of the question and the distribution of student responses across performance levels.

Across dimensions, we observe that CCC assessments generally benefit most from knowledge augmentation, with an average improvement of 13.0\% in accuracy. This aligns with expectations as crosscutting concepts often require broader contextual understanding that can be enhanced through retrieved knowledge. The DCI dimension shows the second-largest improvements, while SEP shows more modest gains, suggesting that scientific practices may be more challenging to evaluate through retrieved content alone.


\subsection{Ablation Study}

In this section, we conduct two ablation experiments. First, to investigate whether the proposed GradeRAG is compatible with in-context learning (ICL), we conducted ablation experiments comparing NonRAG-ICL and GradeRAG-ICL across different shot settings. 
Tables~\ref{tab:ablation_ICL} presents our findings with 3-shot and 6-shot settings.

\vspace{-0.15in}
\paragraph{Effect of Retrieved Knowledge Across Shot Settings}

Comparing across shot settings reveals several patterns. In the zero-shot scenario, adding retrieved knowledge consistently improves performance across nearly all dimensions and questions. For 3-shot learning, the addition of knowledge retrieval also provides benefits, with GradeRAG-ICL outperforming NonRAG-ICL in 7 out of 9 dimension-question combinations. For DCI in both $Q_1$ and $Q_2$, retrieval provides no additional benefit, suggesting the examples may already contain sufficient domain knowledge. With 6-shot learning, GradeRAG-ICL outperforms NonRAG-ICL in only 5 out of 9 cases. For $Q_3$, knowledge retrieval provides no accuracy improvement in any dimension, even with a minor drop in CCC. This suggests that with sufficient examples, additional knowledge becomes redundant for this particular question.
\vspace{-0.15in}

\begin{figure}[htbp]
\Description{Ablation Study of The Retrieval Size}  
\centering
\begin{minipage}[b]{0.475\textwidth}
    \centering
    \includegraphics[width=\textwidth]{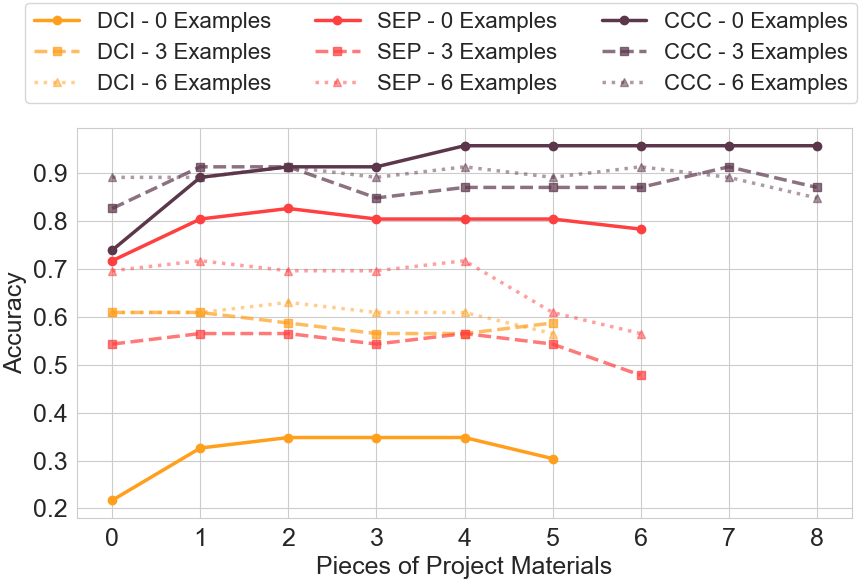}
    (a) Retrieval Size Ablation For $Q_1$
\end{minipage}
\hfill
\begin{minipage}[b]{0.475\textwidth}
    \centering
    \includegraphics[width=\textwidth]{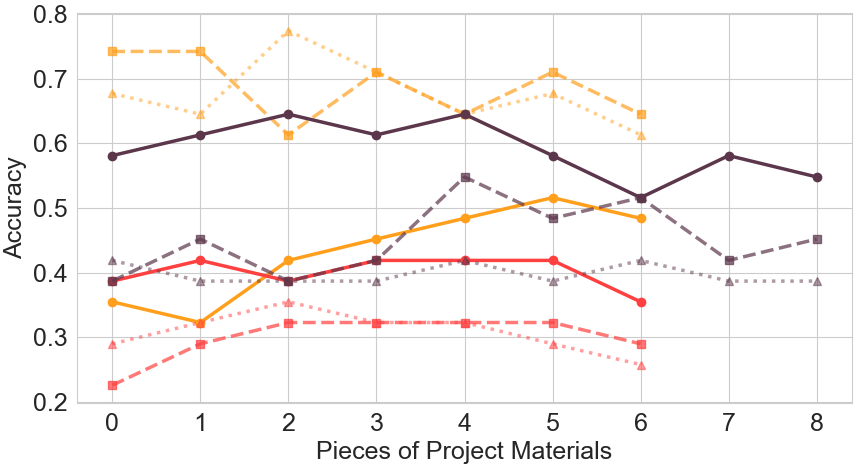}
    (b) Retrieval Size Ablation For $Q_2$
\end{minipage}
\hfill
\begin{minipage}[b]{0.475\textwidth}
    \centering
    \includegraphics[width=\textwidth]{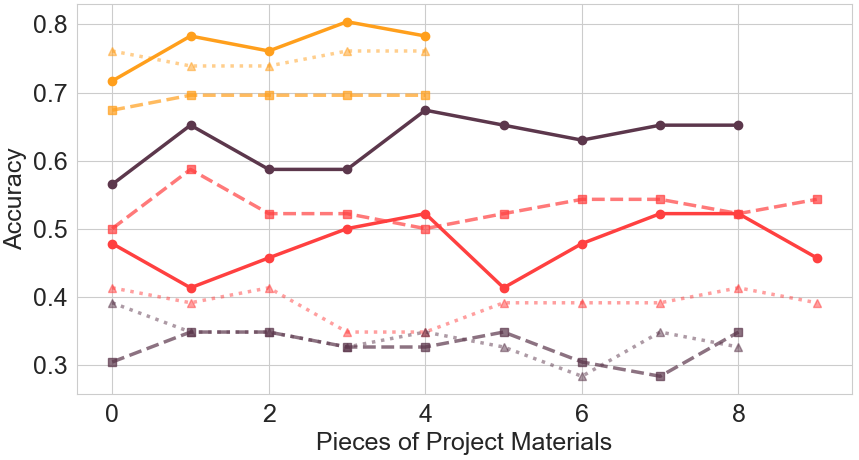}
    (c) Retrieval Size Ablation For $Q_3$
\end{minipage}

\caption{ The impact of the retrieval size ($k$).}
\label{fig:ablation_k}
\end{figure}

\paragraph{Effect of Increasing Example Count}

For NonRAG, increasing examples from 0 to 3 shots substantially improves performance in most cases, particularly for DCI dimensions. However, further increasing from 3 to 6 shots yields miscellaneous results, as we observe some dimensions improve while others deteriorate. For GradeRAG, the impact of increasing examples follows a similar pattern. The transitions from zero-shot to 3-shot generally improve performance, yet the magnitude of improvement is typically smaller than for NonRAG. This suggests that retrieval partially compensates for the lack of examples. The transitions from 3 to 6 shots within GradeRAG show varying effects. Positive gains appear in all questions in DCI and 2 out of 3 in SEP, yet performance decreases for 1 out of 3 in both SEP and CCC.

These two observations suggest the following findings regarding the compatibility of GradeRAG with in-context learning. First, adding more examples might lead to potential information excessiveness, causing diminishing utility or even performance decreases. Second, the optimal balance between retrieval and examples depends heavily on the specific dimension and question complexity. Third, simply adding more examples or more retrieved content does not guarantee improved performance. Different dimensions require different balances between the retrieval size and example number to achieve optimal performance.

\vspace{-0.15in}
\paragraph{Effect of Knowledge Retrieval Size}
We further conducted a systematic ablation study on the impact of retrieval size $k$ (number of retrieved knowledge chunks) across different in-context learning settings. Figures~\ref{fig:ablation_k} represent the accuracy results for all questions across all dimensions as we vary the retrieval size $k$ from 0 to the maximum available chunks. For $Q_1$, increasing $k$ in zero-shot settings generally improves zero-shot performance across all dimensions, with CCC showing the most drastic gains. For $Q_2$, performance typically peaks at moderate $k$ values ($k$=2 or 3) before beginning to decline. $Q_3$ shows more results, with DCI benefiting from increasing $k$ up to 3 in zero-shot settings, while SEP achieves the best performance at $k$=1 under 3-shot settings and CCC shows optimal results with moderate $k$ values in zero-shot settings. Across all three questions, we observe that the optimal retrieval size $k$ decreases as more examples are provided. In zero-shot scenario, $k$=4 generally leads to the best performance; while $k$=1 typically performs the best in 3-shot scenarios and $k$=2 the best in 6-shot scenarios. This suggests that complementary information can be provided by retrieved knowledge or in-context examples.

\section{Conclusion}

This study investigates into the effectiveness of GradeRAG, a novel framework that integrates retrieval-augmented generation into automatic short answer grading for science education. Our experiments across three science assessment tasks demonstrate that incorporating domain-specific knowledge through RAG significantly improves scoring accuracy, with consistent performance gains across multiple dimensions of scientific understanding. 
Our experimental results show several advantages of GradeRAG. First, it effectively enhances the alignment between expert graders and automated grading systems by providing access to relevant domain knowledge. Second, our dual-index strategy and the comprehensive retrieval mechanism ensure that the most pertinent information is retrieved for student responses. Lastly, GradeRAG demonstrates compatibility with in-context learning, exhibiting the complementary relationship between the retrieved knowledge and expert examples, which further enhances grading performance.

\section{Acknowledgments}
This study is supported by the National Science Foundation [grant number DRL-2446701]. Any opinions, findings, and conclusions or recommendations expressed in this material are those of the authors and do not necessarily reflect the views of the National Science Foundation.

%
\bibliographystyle{abbrv}
\bibliography{a_ref}  
%

\end{document}